\documentclass[11pt]{article}

\usepackage[final]{acl}

\usepackage{graphicx}                                        
\usepackage{epstopdf}                                
\usepackage{color}  

\usepackage{times}
\usepackage{latexsym}
\usepackage{enumitem} 
\usepackage{enumerate}

\usepackage[T1]{fontenc}

\usepackage[utf8]{inputenc}

\usepackage{microtype}

\usepackage{inconsolata}

\usepackage{inconsolata}
\usepackage{amsmath}
\usepackage{booktabs}
\usepackage{multirow}
\usepackage{epsfig}
\usepackage{caption}
\usepackage{subcaption}
\usepackage{amsfonts}
\usepackage{algorithm}
\usepackage{algorithmic}
\usepackage{amssymb}
\usepackage{pifont}
\usepackage{xcolor}
\usepackage{tabularx}
\usepackage{bbding}
\usepackage{utfsym}
\usepackage{hyperref}
\usepackage{enumitem}
\usepackage{setspace}
\usepackage{courier} 
\usepackage{url}            
\usepackage{nicefrac}       
\usepackage{microtype}      
\usepackage[edges]{forest}
\usepackage{makecell}
\usepackage{multirow, booktabs, makecell, array} 
\usepackage{tikz}
\usepackage{bbding}
\usepackage{booktabs}

\title{A Survey on Training-free Alignment of Large Language Models}

\author{Birong Pan$^{1}$,  Yongqi Li$^{1}$, Weiyu Zhang$^{2,}$\thanks{\, Corresponding authors.}, Wenpeng Lu$^{2,*}$, \\
{\bf Mayi Xu$^1$, Shen Zhou$^1$, Yuanyuan Zhu$^1$, Ming Zhong$^1$, Tieyun Qian$^{1,3,*}$}\\
        $^1$School of Computer Science, Wuhan University, Wuhan, China \\
        $^2$Faculty of Computer Science and Technology, Qilu University of Technology, Shandong, China \\
        $^3$Zhongguancun Academy, Beijing, China \\
        \texttt{\{panbirong,qty\}@whu.edu.cn} \\}

\begin{document}
\maketitle
\begin{abstract} \label{sec_abstract}

The alignment of large language models (LLMs) aims to ensure their outputs adhere to human values, ethical standards, and legal norms. Traditional alignment methods often rely on resource-intensive fine-tuning (FT), which may suffer from knowledge degradation and face challenges in scenarios where the model accessibility or computational resources are constrained. In contrast, training-free (TF) alignment techniques--leveraging in-context learning, decoding-time adjustments, and post-generation corrections--offer a promising alternative by enabling alignment without heavily retraining LLMs, making them adaptable to both open-source and closed-source environments. This paper presents the first systematic review of TF alignment methods, categorizing them by stages of \textbf{pre-decoding}, \textbf{in-decoding}, and \textbf{post-decoding}. For each stage, we provide a detailed examination from the viewpoint of LLMs and multimodal LLMs (MLLMs), highlighting their mechanisms and limitations. Furthermore, we identify key challenges and future directions, paving the way for more inclusive and effective TF alignment techniques. By synthesizing and organizing the rapidly growing body of research, this survey offers a guidance for practitioners and advances the development of safer and more reliable LLMs.

\end{abstract}

\section{Introduction} \label{sec:introduction}

The advent of Large Language Models (LLMs) \citep{hurst2024gpt, achiam2023gpt, touvron2023llama, chiang2023vicuna} and Multimodal Large Language Models (MLLMs) \citep{bai2023qwen, chen2025sharegpt4v, ye2024mplug} has marked a paradigm shift in human-machine interaction, enabling tasks ranging from complex reasoning to cross-modal understanding. 

\definecolor{hidden-blue}{RGB}{194,232,247}
\definecolor{hidden-black}{RGB}{20,68,106}

\tikzstyle{my-box}=[
    rectangle,
    draw=hidden-black,
    rounded corners,
    text opacity=1,
    minimum height=1.5em,
    minimum width=5em,
    inner sep=2pt,
    align=center,
    fill opacity=.5,
]
\tikzstyle{leaf}=[
    my-box, 
    minimum height=1.5em,
    fill=hidden-blue!90, 
    text=black,
    align=left,
    font=\normalsize,
    inner xsep=2pt,
    inner ysep=4pt,
]
\begin{figure*}[t]
    \vspace{-2mm}
    \centering
    \resizebox{\textwidth}{!}{
        \begin{forest}
            forked edges,
            for tree={
                child anchor=west,
                parent anchor=east,
                grow'=east,
                anchor=west,
                base=left,
                font=\large,
                rectangle,
                draw=hidden-black,
                rounded corners,
                align=left,
                minimum width=4em,
                edge+={darkgray, line width=1pt},
                s sep=3pt,
                inner xsep=2pt,
                inner ysep=3pt,
                line width=0.8pt,
                ver/.style={rotate=90, child anchor=north, parent anchor=south, anchor=center},
            },
            where level=1{text width=7.5em,font=\normalsize,}{},
            where level=2{text width=8em,font=\normalsize,}{},
            where level=3{text width=12em,font=\normalsize,}{},
            [
                TF Alignment Methodologies~(\S\ref{sec_methods}), ver
                [
                    Pre-Decoding TF \\ Alignment~(\S\ref{sec_methods_pre})
                    [
                        Simple Prompt \\ Engineering
                        [
                            URIAL~\cite{lin2023unlocking}{,}
                            AdaShield ~\cite{wang2024adashieldsafeguardingmultimodallarge}{,}\\
                            ICD~\cite{wei2024jailbreakguardalignedlanguage}{,}
                            Anthropological Prompting~\cite{alkhamissi-etal-2024-investigating}{,}\\
                            CoSA  \citep{zhang2024controllablesafetyalignmentinferencetime}{,}
                            SKIG~\cite{sel-etal-2024-skin}{,}
                            BBA \citep{zhao-etal-2024-bba}{,}
                            \\
                            RapGuard~\cite{jiang2024rapguardsafeguardingmultimodallarge}{,}
                            Self-Reminders~\cite{xie2023defending}{,}\\
                            ICDPO~\cite{song-etal-2025-instantly}{,}
                            VILMO \citep{duan2024denevil}
                            , leaf, text width=32.5em
                        ]
                    ]
                    [
                        Enhanced Prompt \\ Strategy
                        [
                            OPO~\cite{xu2023alignflyadaptingchatbot}{,}
                            AUTOCAP~\cite{zhang-etal-2024-autocap}{,}\\
                            Prioritization~\cite{zhang-etal-2024-defending}{,}
                            PRETTY~\cite{zhan-etal-2024-prefix}{,}\\
                            BPO~\cite{cheng-etal-2024-black}{,}
                            PEARL \citep{fu-etal-2024-learning}{,}
                            PICA \citep{liu-etal-2024-take}{,}\\
                            MIXALIGN~\cite{zhang-etal-2024-knowledge-alignment}{,}
                            DUAT \citep{huang-etal-2024-aligning}{,}\\
                            P-Aligner \citep{song2025palignerenablingprealignmentlanguage}{,}
                            ARL2~\cite{zhang-etal-2024-arl2}{,}
                            , leaf, text width=32.5em
                        ]
                    ]
                    [
                        Detector-Based
                        [
                            VLMGUARD ~\cite{du2024vlmguarddefendingvlmsmalicious}{,}
                            CIDER~\cite{xu-etal-2024-cross}{,}\\
                            Token Highlighter~\cite{Hu2024TokenHI}{,}
                            HarmAug~\cite{HarmAug_2024}
                            , leaf, text width=32.5em
                        ]
                    ]
                ]
                [
                    In-Decoding TF \\ Alignment~(\S\ref{sec_methods_in})
                    [
                        Hidden States \\Adjustment
                        [
                            CMRM ~\cite{liu2024unravelingmitigatingsafetyalignment}{,}
                            VLM-Guard~\cite{liu2025vlmguard}
                            , leaf, text width=32.5em
                        ]
                    ]
                    [
                        Logits Difference \\Calculation
                        [
                            DeRa~\cite{liu2024decodingtimerealignmentlanguagemodels}{,}
                            InRa~\cite{zhu2025flexiblerealignmentlanguagemodels}{,}
                            MOD \citep{shi2024decoding}{,}
                            \\
                            Linear Alignment \citep{10.5555/3692070.3692657}{,}
                            Proxy Tuning \citep{liu2024tuninglanguagemodelsproxy}{,}\\
                            $\delta$ -UNLEARNING \citep{huang2024offsetunlearninglargelanguage}{,}
                            GOOD \citep{fang2024good}{,}\\
                            CoCA \citep{Gao2024CoCARS}{,}  
                            IVG~\cite{liu-etal-2024-inference}{,}
                            MCA~\cite{fu2024unlockingdecodingtimecontrollabilitygradientfree}{,}\\
                            Linear Alignment \citep{10.5555/3692070.3692657}
                            , leaf, text width=32.5em
                        ]
                    ]
                    [
                        Guidance-Based
                        [
                            InferAligner~\cite{wang-etal-2024-inferaligner}{,}
                            GenARM~\cite{xu2025genarmrewardguidedgeneration}{,}\\
                            Nudging~\cite{fei2024nudginginferencetimealignmentmodel}{,}
                            RDS~\cite{zeng2025rootdefencestrategiesensuring}{,}
                            PAD \citep{chen2024padpersonalizedalignmentllms}{,}\\
                            Trustworthiness~\cite{qian-etal-2024-towards}{,}
                            Chat Vector \citep{huang-etal-2024-chat}{,}\\
                            Category-Specific Steering~\cite{bhattacharjee2024inferencetimecategorywisesafetysteering}{,}\\
                            OPAD~\cite{zhu2025onthefly}{,}
                            SCANS~\cite{cao2025scans}{,}
                            IAR~\cite{li2025internal}{,}\\
                            WSD \citep{song-etal-2025-well}{,}
                            CARDS \citep{li2025cascaderewardsamplingefficient}{,}
                            \\
                            CAVGAN~\cite{li-etal-2025-cavgan}{,}
                            ARGS \citep{Khanov2024ARGSAA}\\
                            , leaf, text width=32.5em
                        ]
                    ]
                    [
                        Dynamic Search \\Strategy
                        [
                            RAIN \citep{li2024rain}{,}
                            DeAL \citep{huang2024dealdecodingtimealignmentlarge}{,}
                            PAS~\cite{zhu2024personalityalignmentlargelanguage}{,}\\
                            TreeBoN~\cite{qiu2024treebonenhancinginferencetimealignment}{,}
                            DARWIN~\cite{hung2024inferencetimealignmentrewardguided}\\
                            , leaf, text width=32.5em
                        ]
                    ]
                ]
                [
                    Post-Decoding TF \\ Alignment~(\S\ref{sec_methods_post})
                    [
                        Filtering/Correcting \\ Outputs
                        [
                            LLM SELF DEFENSE \citep{phute2024llmselfdefenseself}{,}
                            RA-LLM~\cite{cao-etal-2024-defending}{,}\\
                            Aligner \citep{ji2024alignerefficientalignmentlearning}{,}
                            ETA ~\cite{ding2024etaevaluatingaligningsafety}{,}
                            ECSO \citep{10.1007/978-3-031-72643-9_23}{,}\\
                            PRIVQA \citep{chen2023languagemodelsinstructedprotect}{,}
                            DPS \citep{zhou2024defendinglvlmsvisionattacks}
                            , leaf, text width=32.5em
                        ]
                    ]
                ]
            ]
        \end{forest}
    }
    \caption{Taxonomy of training-free (TF) alignment methodologies for LLMs, categorized into pre-decoding, in-decoding, and post-decoding strategies.}
    \label{fig:overview_methods}
    \vspace{-5mm}
\end{figure*}
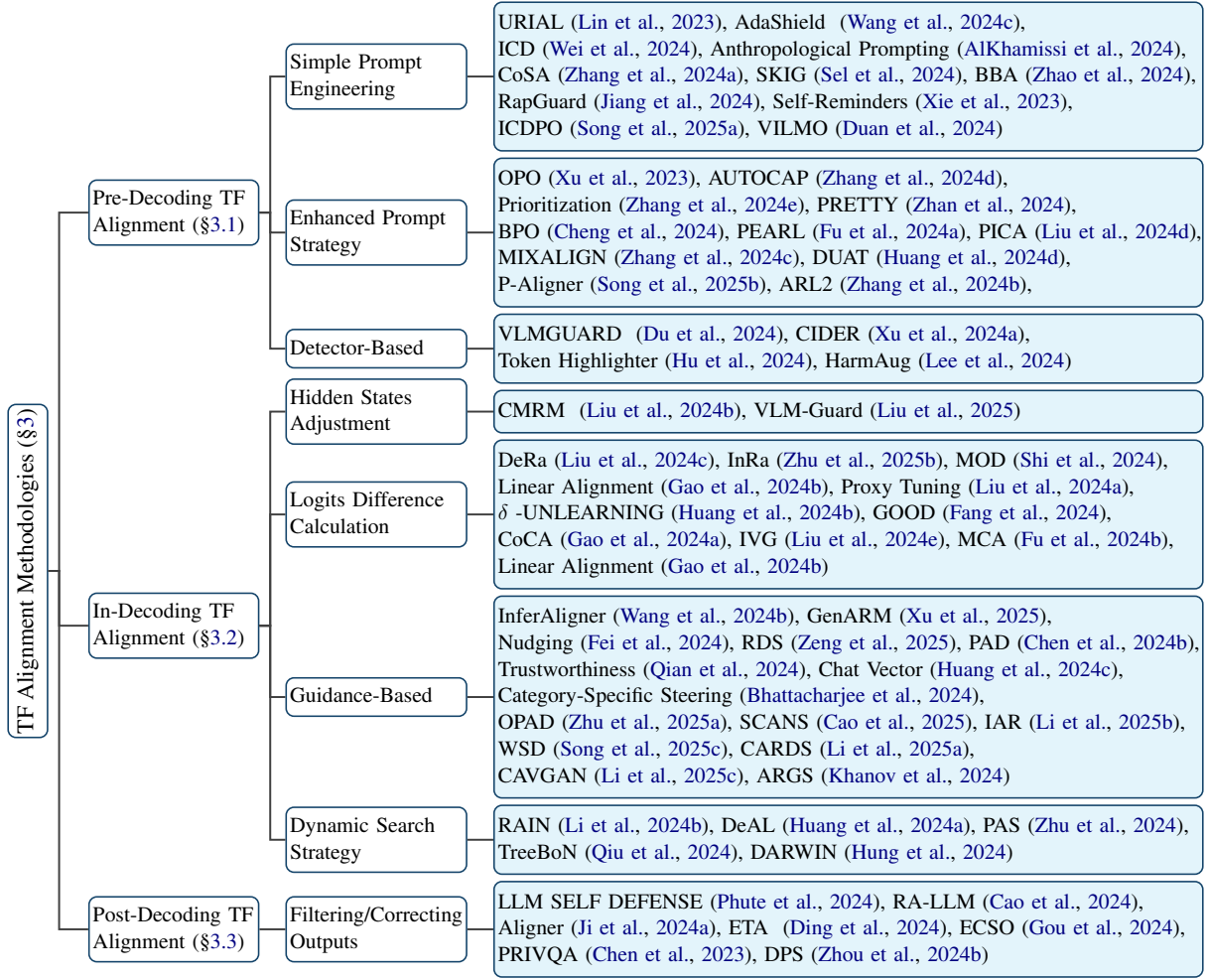

However, as these models permeate critical domains such as healthcare, education, and public discourse, their widespread adoption has ignited a dual-edged societal response: urgent concerns over risks and growing demands for enhanced capabilities. On one hand, LLMs bring about various risks like generating harmful content \citep{soice2023largelanguagemodelsdemocratize}, perpetuating biases and compromising privacy \citep{salecha2024large, staab2024beyond, abdulhai-etal-2024-moral}. On the other hand, users and industries increasingly demand models that adapt to dynamic knowledge \citep{zhang-etal-2023-large, si2023prompting, beukman2023dynamics}, excel in specialized tasks \citep{li-etal-2024-empowering, chen-etal-2024-improving-large}, and deliver personalized services \citep{Richardson2023IntegratingSA, zhuang2024hydra}. These dual imperatives, namely mitigating risks and fulfilling functional demands, underscore the necessity of aligning LLMs with human values, ethical standards, and practical requirements. Alignment is no longer a technique alone but a prerequisite for responsible and effective deployment.

As a critical problem in artificial intelligence and natural language processing, LLM alignment has attracted considerable research attention \citep{shen2023largelanguagemodelalignment, wang2024comprehensivesurveyllmalignment, Gabriel_2020, 10.5555/3600270.3602281}. Traditional alignment methods primarily rely on fine-tuning (FT), where model parameters are adjusted using curated datasets. While effective in specific scenarios, these approaches face one or more of the following three critical limitations: (1) \textit{Severe Knowledge Degradation}: FT may overwrite pretrained knowledge, leading to catastrophic forgetting of general capabilities. (2) \textit{Resource Intensity}: Collecting high-quality alignment data and retraining LLMs requires prohibitive computational and human resources. (3) \textit{Access Constraint}: Proprietary models (e.g., GPT-4, Gemini) restrict parameter access, rendering FT infeasible.

To address these challenges, training-free (TF) alignment has emerged as a versatile paradigm. Instead of modifying model parameters, TF methods intervene at different stages of the generation pipeline: (1) \textit{Pre-Decoding TF Alignment}: guiding models by modifying inputs or prompts. (2) \textit{In-Decoding TF Alignment}: adjusting token selection during generation. (3) \textit{Post-Decoding TF Alignment}: filtering or refining outputs to meet safety and utility criteria. The overview of TF alignment methods at different stages is shown in Figure~\ref{fig:overview_methods}, which helps to gain a comprehensive understanding of the process.

While TF alignment demonstrates notable advantages, our analysis reveals critical challenges and promising further directions, including (1) maintaining general capabilities, (2) exploring cost-effective frameworks, (3) overcoming generalization of TF alignment, (4) developing controllable TF alignment methods.
This survey not only seeks to inspire further research interests in this area but also aims to guide the evolution of LLMs toward benefiting human society.

In summary, the main contributions of this paper are as follows:

\begin{itemize} [leftmargin=*, itemsep=0pt, topsep=0pt]
    \item \textbf{First Review of TF Alignment}: To the best of our knowledge, this is the first comprehensive review of TF alignment methods for LLMs.
    \item \textbf{New Taxonomy Framework}: We present a systematic framework for TF alignment approaches, which organizes and categorizes the existing work in a structured way, covering pre-decoding interventions, in-decoding adjustments, and post-decoding refinement. 
    \item \textbf{Prospective Research Roadmap}: We outline critical and challenging future directions for TF alignment, addressing key areas such as cost-effective framework, and controllable alignment. This roadmap provides a foundation for advancing TF alignment research towards more robust and inclusive LLMs.
\end{itemize}




\section{Preliminary}
\label{sec_concepts}

This section systematically explores four pivotal dimensions of TF alignment, including alignment-related concepts, the rationale for choosing TF alignment, scenarios where TF alignment is preferable, and methods to evaluate TF alignment approaches.

\subsection{What are LLM Alignment and TF Alignment?}
\label{sec_concepts_definition}

LLM alignment is the process of ensuring that language models behave in ways that align with human expectations, societal values, legal standards, and ethical principles. This review focuses on TF alignment, which encompasses functional alignment and normative alignment.

Functional alignment emphasizes the model's ability to perform tasks accurately and reliably. It can be further divided into two categories: (1) \textit{Task-specific Alignment}: Tailoring the model for specific applications (e.g., \citet{zhan-etal-2024-prefix}) to optimize the objective towards particular requirements. (2) \textit{General Capability Alignment}: Ensuring the model maintains updated knowledge and strong generalization abilities across diverse tasks and scenarios (e.g., \citet{zhang-etal-2024-knowledge-alignment}).

Normative alignment focuses on ethical, safety, and user-centric aspects. It also contains two main components: (1) \textit{Public Safety and Ethical Alignment}: Ensuring model outputs comply with safety standards, ethical norms, and regulatory guidelines to avoid harmful content and data privacy violations (e.g., \citet{fang2024good}). The goal is to mitigate potential risks and prevent adverse impacts on society. (2) \textit{Personalized Alignment}: Adapting the model’s interaction style and outputs to individual preferences and usage patterns, balancing safety protocols with personalized experiences (e.g., \citet{chen2024padpersonalizedalignmentllms}).

In conclusion, functional alignment ensures the model performs tasks effectively, while normative alignment guarantees that its operation remains safe and ethical.





\subsection{Why Opt for TF Alignment?} \label{section_concepts_motivation}
Building upon the superficial nature of alignment tuning demonstrated by LIMA \citep{zhou2023lima}, the extended empirical discovery in URIAL \citep{lin2023unlocking} demonstrates that TF alignment can match or even surpass FT-aligned LLM performance. TF alignment technology warrants deeper analysis to advance future LLM research. Therefore, we undertake a systematic investigation into TF alignment.

As listed in introduction, traditional alignment methods based on FT have  one or more critical limitations in practice\footnote{For example, modern adapter-based FT techniques can significantly reduce computational cost and are less energy-intensive, but the data dependency issue and the closed-source model accessibility issue still exist.}. In contrast, TF alignment provides a versatile and efficient alternative. The core advantages of TF alignment are as follows.

\noindent \textbf{Cost Efficiency and Environmental Sustainability} 
(1) \textit{No Training Overheads}: TF methods reduce computational costs (e.g., GPU hours) and hyperparameter tuning burdens since they eliminate the need to train LLMs. (2) \textit{Reduced Data Dependency}: Unlike supervised fine-tuning (SFT), they often require no labeled datasets, minimizing annotation labor and data collection costs. 

\noindent \textbf{Operational Flexibility and Model Compatibility} (1) \textit{Plug-and-Play}: Techniques like in-context learning can be applied ``on the fly'' \citep{min-etal-2022-rethinking}, enabling rapid iteration (e.g., switching alignment objectives without retraining). (2) \textit{Low Storage Overhead}: Unlike parameter-efficient fine-tuning (PEFT) methods (e.g., LoRA \citep{hu2021lora}, TF alignment introduces low additional storage costs. (3) \textit{Black-Box Adaptability}: Most TF alignment methods operate without accessing internal model parameters, making them compatible with both open- and closed-source models (e.g., Llama, GPT-4, Claude).

\noindent \textbf{Slight Knowledge Degradation} By avoiding parameter updates or making minor adjustments, TF alignment methods better preserve the original knowledge of LLMs.

\subsection{How to Evaluate LLM Alignment?} \label{section_concepts_evaluate}
The evaluation of LLM alignment effectiveness can be categorized into three primary paradigms based on assessment methods.

\noindent \textbf{Benchmark-Driven Evaluation} 
This quantitative approach utilizes standardized test suites with predefined metrics to measure alignment performance. For instance, MMLU \citep{hendrycks2020measuring} and XSum~\cite{narayan-etal-2018-dont} benchmarks assess general capability retention after alignment, while perplexity measurements \citep{6773024} evaluate prediction quality by quantifying the cross-entropy between model outputs and reference distributions.

\noindent \textbf{Human-Centric Evaluation} 
This paradigm employs human judgment through two primary modalities: crowdsourcing annotation and expert analysis. Large-scale evaluations often leverage platforms like Amazon Mechanical Turk\footnote{https://www.mturk.com} to collect judgments from workers. Also, domain-specific evaluations can be conducted by experts performing granular behavioral analyses.

\noindent \textbf{Model-Assisted Evaluation} 
Emerging automated methods employ auxiliary LLMs as evaluators through two principal strategies: prompt-based assessment and self-examination techniques. The former implements critic models (e.g., GPT-4) for direct quality scoring via instructional prompting, while the latter verifies alignment consistency through generated explanations that expose model reasoning patterns.

\section{TF Alignment Methods} \label{sec_methods}

\begin{figure*}[ht]
    \centering
    \includegraphics[width=\linewidth]{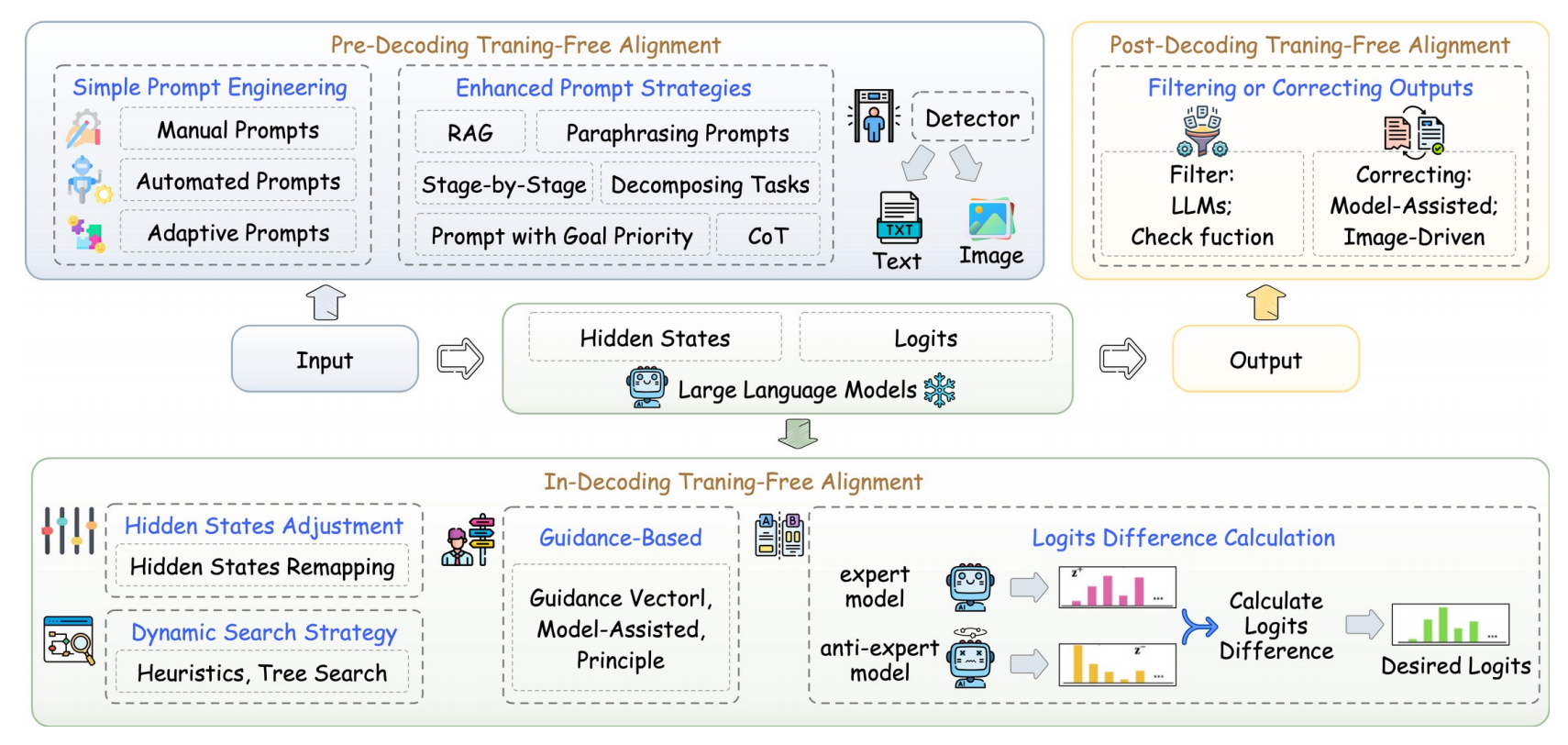}
    \caption {A conceptual framework illustrating training-free (TF) alignment strategies for large language models (LLMs), categorized into pre-decoding, in-decoding, and post-decoding stages.}
    \label{fig:Different Stages of TF Alignment}
\end{figure*}

TF alignment methods can intervene at different stages of the generation pipeline without heavily retraining LLMs. Hence we categorize them into Pre-Decoding, In-Decoding, and Post-Decoding TF alignment, analyzing their application to unimodal LLMs and MLLMs. Figure~\ref{fig:Different Stages of TF Alignment} shows TF alignment methods at different stages, as well as their specific mechanisms.

\subsection{Pre-Decoding TF Alignment} \label{sec_methods_pre}
Pre-decoding TF alignment methods align models by modifying or detecting inputs before decoding without changing models' internal parameters. These approaches are lightweight and especially suitable for black-box LLMs.

\paragraph{Simple Prompt Engineering} One of the simplest yet effective alignment strategies involves crafting specific prompts to elicit desired responses. ICL has emerged as a powerful tool for aligning LLMs with human preferences, a technique known as \textit{In-Context Alignment (ICA)}. For instance, URIAL \citep{lin2023unlocking} demonstrates that just three in-context examples and a single system prompt can effectively align pre-trained LLMs, achieving performance comparable to FT methods while significantly reducing costs. Similarly, Anthropological Prompting \citep{alkhamissi-etal-2024-investigating} and VILMO \citep{duan2024denevil} leverage specific prompts to enhance alignment. Besides, ICDPO \citep{song-etal-2025-instantly} leverages ICL and instant scorer to enhance the final performance.
To tackle the inflexibility of existing alignment methods when confronted with diverse social norms across different cultures and regions, CoSA \citep{zhang2024controllablesafetyalignmentinferencetime} introduces safety configs, which allow models to dynamically adapt during inference based on these configurations, thereby catering to varying requirements.

Although MLLMs introduce a novel visual modality, enabling the embedding of harmful content within images to circumvent security mechanisms, the prompt engineering strategy in pre-decoding remains applicable. To defend against structured-based jailbreak attacks, AdaShield \citep{wang2024adashieldsafeguardingmultimodallarge} includes both manual static and adaptive defense prompts, the latter being iteratively optimized through collaboration between the target MLLM and an LLM-based defense prompt generator. Similarly, to address the limitation that static safety guidelines fail to account for specific risks in different multimodal contexts, ICD \citep{wei2024jailbreakguardalignedlanguage}, RapGuard \citep{jiang2024rapguardsafeguardingmultimodallarge}, and BBA \citep{zhao-etal-2024-bba} generate scenario-specific security prompts. They leverage multimodal reasoning to dynamically identify and mitigate risks.

Prompts can also act as self-reminders \citep{xie2023defending}, packaging user queries and prompting ChatGPT to respond responsibly. SKIG \citep{sel-etal-2024-skin} designs multi-dimensional prompts that stimulate the model's sense of responsibility, exploring the consequences of decisions from multiple stakeholders' perspectives. This approach enhances the model’s moral reasoning ability. 

\paragraph{Enhanced Prompt Strategy} Beyond simple prompts, this category encompasses methods that enhance prompts through additional techniques or iterative refinement. Certain human values to be aligned often vary with time and place, OPO \citep{xu2023alignflyadaptingchatbot} uses retrieval-augmented-generation (RAG) can address the ever-changing nature of alignment in human values. AUTOCAP \citep{zhang-etal-2024-autocap} integrates Chain-of-Thought (CoT) reasoning paths across languages to improve cross-lingual alignment \citep{shi2022languagemodelsmultilingualchainofthought, Tanwar2023MultilingualLA, qin-etal-2023-cross}. 
P-Aligner \citep{song2025palignerenablingprealignmentlanguage} automatically rewrites user instructions with pre-defined principles, significantly boosting downstream LLM alignment.
Other methods focus on feeding inputs in a more structured way. For example, \citet{zhang-etal-2024-defending} introduce goal priority through few-shot prompting, instructing the LLM to prioritize safety over helpfulness. PRETTY \citep{zhan-etal-2024-prefix} improves alignment by adding task-related prior markers to the input prefix, narrowing the performance gap between TF models and FT models. BPO \citep{cheng-etal-2024-black} optimizes user prompts to suit LLMs’ input understanding, ensuring user intents are realized without updating internal parameters. PEARL \citep{fu-etal-2024-learning} paraphrases questions in expressions preferred by the model, ensuring better alignment with user expectations. 

Additionally, the decomposition of complex problems into subproblems has proven effective. For example, MIXALIGN \citep{zhang-etal-2024-knowledge-alignment} and DUAT \citep{huang-etal-2024-aligning} break down difficult problems into subtasks and designs corresponding prompts step-by-step, fully utilizing the generation and reasoning abilities of LLMs to dynamically solve knowledge alignment challenges. Another way uses prompts at the first stage, followed by additional operations to further refine the output. PICA \citep{liu-etal-2024-take} operates in two stages: first, the model generates prior response tokens via ICL while extracting an ICL vector; second, this ICL vector guides the model to generate desired responses without requiring further demonstrations. This approach ensures consistent alignment while minimizing the need for extensive prompting.

\paragraph{Detector-Based} 
VLMGUARD \citep{du2024vlmguarddefendingvlmsmalicious}  utilizes unlabeled user prompt data for malicious prompts detection, effectively distinguishing between malicious and benign samples without additional manual annotation. CIDER \citep{xu-etal-2024-cross} detects malicious image inputs by analyzing the semantic similarity between text and image modalities. Token Highlighter~\cite{Hu2024TokenHI} locates the jailbreak-critical tokens and use soft removal technique to align.
To reduce the cost (e.g., substantial memory requirements and latency) of detector with billions of parameters, HarmAug~\cite{HarmAug_2024} distills a large teacher safety guard model into a smaller one.

\paragraph{Key Insights} 

Pre-decoding methods aim to reformulate queries, append exemplars, or leverage detectors before feeding inputs into LLMs, ensuring model outputs align with human standards. These approaches offer advantages of simplicity and broad applicability, working seamlessly with both open-source and closed-source models. However, they suffer from generalization limitations due to reliance on few-shot examples or manual prompt engineering. For instance, CoSA \citep{zhang2024controllablesafetyalignmentinferencetime} introduces safety configurations to enable dynamic inferential adaptation, yet this only addresses safety scenarios—cross-scenario generalization, cultural variance, and model-agnostic adaptability remain underexplored. Additionally, most prompt engineering techniques are constrained by context window limits (e.g., LLaMA-2-13B's 4096-token boundary \citep{machlab2024llmincontextrecallprompt}) or suffer semantic drift \citep{10.1145/3571730} due to excessively long inputs. The subsequent in-decoding and post-decoding methods tackle these limitations from alternative perspectives.

\subsection{In-Decoding TF Alignment} \label{sec_methods_in}
Aligning LLMs before decoding often appears ideal following the principle of ``an ounce of prevention is worth a pound of cure''. However, not all undesired behavior can be fully addressed at this stage. This motivates in-decoding TF alignment, which adjusts model behaviors during decoding. These methods typically modify hidden states or logits, through remapping hidden states, calculating logits difference, exacting guidance signals, or employing search strategies. 

\paragraph{Hidden States Adjustment}
The disjoint training protocols between visual and linguistic modules in VLMs induce latent space fragmentation, where their representations form distinct cluster regions with significant distributional discrepancy, manifesting diminished safety-aligned capabilities. To address these issues, CMRM \citep{liu2024unravelingmitigatingsafetyalignment} and VLM-Guard~\cite{liu2025vlmguard} mitigate this by pulling hidden states back into the representation space optimized by LLMs, thus restoring secure alignment capabilities.

\paragraph{Logits Difference Calculation} 
Several methods modify the logits of tokens during generation to align outputs with desired behaviors. 
\cite{liu2024decodingtimerealignmentlanguagemodels, liu2024tuninglanguagemodelsproxy, shi2024decoding, fang2024good, zhu2025flexiblerealignmentlanguagemodels} combine the logits of the alignment model and a reference model to guide generation, achieving FT-like effects without parameter updates. 
$\delta$ -UNLEARNING \citep{huang2024offsetunlearninglargelanguage} learns an offset from logits comparisons between smaller models, applying it to larger black-box LLMs to adjust their predictive behavior. 
Additionally, CoCA \citep{Gao2024CoCARS} calibrates the output distribution by amplifying the model’s response to safety prompts. The method calculates logits difference before and after safety prompt insertion, ensuring robust alignment. 
Recently, IVG~\cite{liu-etal-2024-inference} uses implicit and explicit value functions to guide language model decoding at token and chunk-level respectively, efficiently aligning LLMs purely at inference time.
Contrastive decoding has also been adopted. For example, MCA \citep{fu2024unlockingdecodingtimecontrollabilitygradientfree} constructs expert and adversarial prompts to promote or suppress aligned targets dynamically. 
Lately, Linear Alignment \citep{10.5555/3692070.3692657} relies on a novel parameterization for policy optimization under divergence constraints and estimates the preference direction using self-contrastive decoding.

\paragraph{Guidance-Based}
For guidance-based approaches, \cite{qian-etal-2024-towards, wang-etal-2024-inferaligner, huang-etal-2024-chat, bhattacharjee2024inferencetimecategorywisesafetysteering, cao2025scans, li2025internal} extract guidance vectors to adjust the activation states of the target model during inference, achieving alignment. 
To predict next-token rewards for efficient and effective autoregressive generation,  GenARM~\cite{xu2025genarmrewardguidedgeneration} and CARDS \citep{li2025cascaderewardsamplingefficient} leverage the reward model to guide frozen LLMs toward expected distribution.
PAD \citep{chen2024padpersonalizedalignmentllms} and ARGS \citep{Khanov2024ARGSAA} leverage token-level rewards to guide the decoding process, dynamically guiding the base model’s predictions.
Nudging~\cite{fei2024nudginginferencetimealignmentmodel} and WSD \citep{song-etal-2025-well} combine a large base model with a much smaller aligned model.
CAVGAN~\cite{li-etal-2025-cavgan} utilizes generative adversarial network to learn the security judgment boundary inside the LLM to achieve efficient alignment. 
RDS~\cite{zeng2025rootdefencestrategiesensuring} designs a root classifier based on the discriminative capacity of queries, and then reorders the token and prioritizes the benign token. 
To directly align model outputs with human preferences, OPAD~\cite{zhu2025onthefly} guides the generation of a final output that aligns with the target principle in a token-by-token manner.

\paragraph{Dynamic Search Strategy} 
Search strategies have also been frequently used for alignment. RAIN \citep{li2024rain} mirrors human behavioral patterns, which allows LLMs to evaluate their own generation and use the evaluation results to guide rewind and generation for self-alignment. 
Similarly, DeAL \citep{huang2024dealdecodingtimealignmentlarge} views decoding as a heuristic-guided search process and facilitates the use of a wide variety of alignment objectives. TreeBoN \citep{qiu2024treebonenhancinginferencetimealignment} combines tree search strategies with Best-of-N (BoN) sampling, iteratively expanding high-quality partial responses while pruning low-quality candidates to improve alignment quality and efficiency. DARWIN \citep{hung2024inferencetimealignmentrewardguided} achieves the right balance between exploration and exploitation of rewards during decoding with evolutionary heuristics. Additionally, \citet{zhu2024personalityalignmentlargelanguage} proposed PAS, which includes direction search and distance search. They focuse on the alignment with personal traits and develop an activation intervention optimization method to enhance LLMs’ ability to efficiently align with individual behavioral preferences using minimal data and computational resources.

\paragraph{Key Insights} 

In-decoding TF alignment methods enable dynamic, fine-grained output control via hidden state adjustment, logit modification, tree search strategies, etc. For MLLMs, projecting hidden states into LLM-optimized representation spaces shows promise in resolving multimodal integration challenges. While highly effective for alignment and alleviating certain generalization issues, this paradigm often requires access to model internals, restricting applicability in black-box systems. Moreover, techniques like Trustworthiness and Category-Specific Steering \citep{qian-etal-2024-towards, bhattacharjee2024inferencetimecategorywisesafetysteering}, which modify hidden or activation states, can inadvertently compromise the model's pre-existing task-specific knowledge. These trade-offs directly inform the research directions outlined in Section \ref{section_future}.


\begin{table*}[]
\centering
\resizebox{0.8\textwidth}{!}{\begin{tabular}{@{}ccccc@{}}
\toprule
\textbf{Model}    & \textbf{Type}      & \textbf{AdvBench $\uparrow$} & \textbf{SafeEdit $\uparrow$} & \textbf{TruthfulQA $\downarrow$} \\ \midrule
llama2-7b-chat & Defaults & 99.78     & 37.60     & 5.05                                  \\
SafeDecoding  & FT Alignment & 100.00    & 94.60     & 54.44                                                        \\
URIAL    & Pre-Decoding TF Alignment     & 99.34     & 66.60     & 15.94                                   \\
SCANS   & In-Decoding TF Alignment      & 99.34     & 97.80     & 0.80                                   \\
RA-LLM  & Post-Decoding TF Alignment       & 100.00    & 98.00     & 36.12                                          \\ \bottomrule
\end{tabular}}

\caption{\label{tab:numerical2}
The numerical  comparison of  the performance of llama2-7b-chat with different safety alignment methods when facing with malicious questions, adversarial attacks and safe questions. 
On AdvBench and SafeEdit datasets, numerical values denote the Defense Success Rate, where $\uparrow$ indicates that higher values are preferable. On TruthfulQA, the numerical values represent the Benign Refusing Rate, and $\downarrow$ signifies that lower values are more desirable.
}
\vspace{-5mm}
\end{table*}

\subsection{Post-Decoding TF Alignment} \label{sec_methods_post}
In-decoding TF alignment methods often require access to token logits and vocabulary, restricting their applicability to models from the same series. To overcome these limitations, post-decoding TF alignment methods are proposed, which operate on the generated outputs without accessing the model internals.

Post-decoding TF alignment methods often adopt strategies to filter or correct outputs. \citet{phute2024llmselfdefenseself} utilize inherent language comprehension of LLMs to self-examine generated text. RA-LLM \citep{cao-etal-2024-defending} uses an alignment check function, enhancing robustness by randomly discarding portions of the input request to mitigate adversarial perturbations. For correction-based approaches, Aligner \citep{ji2024alignerefficientalignmentlearning} trains a separate model to learn the residual between the initial and aligned outputs. This plug-and-play method redistributes initial answers into more helpful and harmless responses.

Multimodal applications introduce additional challenges, such as visual inputs that may contain toxicity. ETA \citep{ding2024etaevaluatingaligningsafety} uses a multimodal evaluator to assess visual inputs with the CLIP score. If need to align, ETA can operate shallow alignment (interference prefix) and deep alignment (sentence-level best-of-N searching). ECSO \citep{10.1007/978-3-031-72643-9_23} also reviews responses first. If deemed unsafe, it converts images into text via query-aware transformation, enabling the pre-aligned LLM to handle both text-based and image-embedded malicious content for safe output generation. \citet{chen2023languagemodelsinstructedprotect} propose an instruction based on self-moderation, deciding whether to respond based on privacy concerns. DPS \citep{zhou2024defendinglvlmsvisionattacks} integrates LLM security checkers to filter responses, defending against visual attacks while maintaining clean input performance. This reduces harmful content while preserving model performance.

\paragraph{Key Insights} 
Analogous to pre-decoding approaches, post-decoding TF alignment methods are model agnostic. Also, they mitigate the generalization constraints seen in pre-decoding stages, offering heightened versatility. In MLLM contexts, integrating visual-textual data enables dynamic defense mechanisms, though this amplifies alignment complexity. A notable trade-off is the latency introduced by filtering or correction processes.




\subsection{Quantitative Analysis}
To have an insight into diverse alignment methods, we conduct experiments on llama2-7b-chat with safety alignment for illustration.

\noindent \textbf{Datasets}
Following SCANS~\cite{cao2025scans}, we select AdvBench and TruthfulQA~\cite{Lin2021TruthfulQAMH} datasets to evaluate the safety and helpfulness of alignment methods. Building on this, we incorporate a more challenging dataset, SafeEdit~\cite{wang-etal-2024-detoxifying}, to assess how the aligned models perform against adversarial attacks.

\noindent \textbf{Methods} 
For FT alignment methods, we select SafeDecoding~\cite{xu2024safedecoding}, a relatively new method that accounts for jailbreak attacks and demonstrates superior performance. For TF alignment methods, we select one recent approach for each of three stages~\cite{lin2023unlocking, cao2025scans, cao-etal-2024-defending}.

\noindent \textbf{Results}
The results are shown in Table~\ref{tab:numerical2}. Notably, TF alignment methods can match or even exceed the safety and helpfulness performance of FT alignment methods. For example, SCANS outperforms SafeDecoding on SafeEdit and TruthfulQA datasets. It also shows performance comparable to SafeDecoding on AdvBench.
This reflects the effectiveness of TF alignment as a supplement to FT alignment, demonstrating that TF methods merit deeper investigation, particularly in scenarios where FT alignment faces practical constraints (e.g., incompatibility with closed-source models, significant knowledge degradation during fine-tuning). However, TF alignment methods also exhibit certain limitations. For example, TF methods like URIAL and RA-LLM compromise the model's helpfulness, which certainly damages the helpful performance on TruthfulQA, albeit such damage is less severe compared to the FT method SafeDecoding. More challenges will be described in the next section. 
We also analyze practical deployment considerations for different TF alignment methods in Table~\ref{tab:comparison2} of Appendix~\ref{sec:appendix}.

Please refer to Appendix~\ref{sec:appendix} for more details on datasets, evaluation criteria, result analysis, and comparison of different TF alignment methods. 

\section{Discussion} \label{sec_Discussion}
While the above TF alignment methods demonstrate notable advantages, significant challenges still exist. Meanwhile, there are many opportunities to make TF alignment more effective, efficient, and adaptive.

\subsection{Open Challenges} \label{sec_Discussion_1}
While TF alignment offers practical advantages, our systematic analysis reveals three fundamental challenges that constrain its broader adoption: (1) \textit{Degradation of General Capabilities}: While most TF alignment methods preserve model knowledge integrity better than FT alignment methods, certain in-decoding methods \citep{qian-etal-2024-towards, bhattacharjee2024inferencetimecategorywisesafetysteering} potentially compromise general performance. Because these methods adjust the hidden states or activation states of the target model during inference, such operations undermine the model's inherent knowledge for other tasks. This interference disrupts the pre-trained model's generalized ability to handle diverse tasks, as the adjusted states may no longer align with the optimal representations learned for cross-task scenarios. Consequently, the model's performance in non-target tasks might deteriorate, highlighting the trade-off between task-specific adaptation and maintenance of general capabilities in alignment strategies.
(2) \textit{Increased Inference Overhead}: TF alignment approaches often introduce additional latency. Whether through input detection, logits adjustment, or modification of generated responses, the additional computational steps result in higher latency. 
(3) \textit{Difficult to Generalize}: TF alignment cannot leverage extensive alignment data. As a result, it may fail to generalize effectively to unseen or more challenging scenarios (e.g. adversarial attacks), as it lacks exposure to diverse or complex cases.


\subsection{Future Directions}  \label{section_future}
\paragraph{Maintain General Capabilities}
Develop mechanisms to maintain or enhance the general capabilities of LLMs while performing alignment, such as incorporating auxiliary loss functions that balance alignment objectives with model performance.
Another promising way is exploring lightweight interventions in hidden states or logits that minimize disruption to the model's original outputs, leveraging techniques like sparse interventions. For instance, when adjusting hidden states, task-relevant neurons \cite{leng2025understandingmultitasklearninggeneralization, tang-etal-2024-language} can be identified, eliminating the need to modify the entire state space and thereby preserving a significant portion of the original performance.

\paragraph{Explore Unified Metrics and Cost-Effective Frameworks}
To compare various alignment methods more thoroughly, unified computational overhead metrics to evaluate memory usage, CPU/GPU utilization, and time taken for alignment steps need to be established. Additionally, cost-effective algorithms for input detection, logits adjustment, and response modification to reduce latency should be designed. For instance, HarmAug~\citep{HarmAug_2024} distills a large teacher safety guard model into a smaller one. 
Also, it is necessary to investigate on-the-fly optimization strategies \cite{zhu2025onthefly} that dynamically adapt computational resources based on complexity of tasks or efficiency requirements of users.

\paragraph{Overcome Insufficient Generalization}
When addressing the diverse scenarios of different cultures, applications, or users, it is essential to establish corresponding principles for guidance. For instance, CoSA \citep{zhang2024controllablesafetyalignmentinferencetime} introduces safety configs, which allow models to dynamically adapt during inference based on these configurations, thereby catering to varying requirements. However, this only accounts for different safety scenarios, while the generalization across other scenarios and diverse cultures, as well as the adaptability of various models, remains equally worthy of exploration.
Additionally, incorporating counterfactual samples into prompts can break spurious correlations, compelling the model to learn causal features and thereby enhancing its generalization capability.
Meta-learning \cite{finn2017model,khoee2024domain} techniques can also be leveraged, enabling TF alignment methods to quickly adapt to new, unseen scenarios with minimal additional data. Furthermore, explore hybrid approaches that combine the strengths of FT (rich alignment data) and TF alignment (no parameter updates) to achieve better generalization.


\paragraph{TF Alignment for Uni-Modal Model}


Although some efforts have been made in the field of TF alignment for MLLMs, they are currently limited to scenarios where the input is multimodal (e.g., image and text) and the output is purely textual. Extending alignment to cases where the output includes multimodal content, such as images, introduces new challenges and complexities \cite{xiong2024autoregressivemodelsvisionsurvey, chameleonteam2024chameleonmixedmodalearlyfusionfoundation, xie2024showosingletransformerunify, zhou2024transfusionpredicttokendiffuse}, making it a promising and worthwhile direction for future research. 
\citet{ji2024alignanythingtrainingallmodality} have made initial strides in aligning all-modality models with human intentions through FT methods. 
However, research on TF alignment for such Uni-Modal Models is still lacking.

\paragraph{Controllable TF Alignment}
Interpretability can help address this problem \citep{wu2024usablexai10strategies}. It locates and edits fine-grained features, which can be used to align LLMs to human values.
For example, by feeding prompts with negative and positive prefixes into LLMs, \citet{leong2023selfdetoxifyinglanguagemodelstoxification} analyze internal contextualized representations to identify the toxicity direction of each attention head.
However, many interpretability studies are primarily based on toy or theoretical models. Developing more practical interpretability TF alignment methods is necessary.
Another way is to use principle-based guidance. OPAD~\cite{zhu2025onthefly} designs a principle-guided reward function to align model outputs with human preferences without FT. Moreover, considering the advantages of TF alignment in customization, explore the use of controllable TF alignment for personalized scenarios \cite{zhu2025onthefly, chen2024padpersonalizedalignmentllms}.

\section{Conclusion} \label{sec_conclusion}


In this paper, we present a comprehensive survey of training-free (TF) alignment methods for LLMs, making three significant contributions to the field. First, we establish a conceptual framework for the TF alignment. Second, we propose a novel taxonomy that categorizes TF alignment techniques into three distinct paradigms: pre-decoding interventions, in-decoding adjustments, and post-decoding refinement strategies. Finally, we identify promising research directions, emphasizing the need for effective, efficient, and adaptive TF alignment methods.

This survey is expected to serve both as an introductory guide for newcomers and a reference manual for practitioners, aiming to accelerate progress in developing efficient, ethical, and human-aligned LLM systems. We hope that our investigation will inspire novel research addressing the critical challenges and future directions of TF alignment.

\section*{Limitations}

While this survey provides a systematic overview of TF alignment methods for LLMs, several inherent limitations in its scope need more discussion:
Firstly, in this review, base models must be sufficiently capable of responding effectively to TF alignment techniques, which means that these models might require prior fine-tuning.
Secondly, we pay more attention to safety TF alignment due to page limitation, which makes it infeasible to provide a detailed overview of all TF alignment approaches. Thus, we choose to focus intensively on one category.
Finally, experiments just conducted on llama2-7b-chat because the nature of our literature survey. It is hoped that our preliminary exploration will inspire further researches on TF alignment.

\section*{Ethics Statement}
Our work is entirely at the methodological level, which means that there will not be any negative social impacts. 

\section*{Acknowledgments}
This work was supported by a grant from the National Natural Science Foundation of China (NSFC) project (No. 62276193),  the Key Laboratory of Computing Power Network and Information Security, Ministry of Education under Grant No. 2024ZD027, and the Fundamental Research Funds for the Central Universities, China (No. 2042022dx0001).

\bibliography{main}

\newpage
\appendix

\section{Appendix}  \label{sec:appendix}

\newcommand{\crosscheck}{ 
    \ooalign{
        \raisebox{-0.4ex}{\scriptsize\XSolidBrush}\cr 
        \hidewidth$\CheckmarkBold$\hidewidth\cr 
    }
}

\begin{table*}[htp]
  \centering
  \resizebox{\textwidth}{!}{
    \begin{tabular}{p{7.915em}p{9.5em}p{6.915em}p{7.165em}p{5.665em}p{5.335em}p{7.665em}}
    \toprule
    \multirow{2}[4]{*}{\centering \textbf{Category}} & 
    \multirow{2}[4]{*}{\centering \textbf{Characteristic}} & 
    \multicolumn{2}{p{14.08em}}{\centering \textbf{Compatibility}} & 
    \multicolumn{2}{p{11em}}{\centering \textbf{Efficiency}} & 
    \multirow{2}[4]{*}{\centering \textbf{Generalization}} \\
\cmidrule{3-6}    \multicolumn{1}{c}{} & \multicolumn{1}{c}{} &
\parbox[c]{6.915em}{\centering\textbf{Model\\Accessibility}}  & \textbf{Plug-and-Play} & \parbox[c]{5.665em}{\centering\textbf{Storage\\Efficiency}} & \parbox[c]{5.335em}{\centering\textbf{Real-Time\\Efficiency}}& \multicolumn{1}{c}{} \\
    \midrule
    \multirow{3}[2]{*}{\makecell[c]{Pre-Decoding\\TF Alignment}} & \makecell[c]{Simple Prompt\\Engineering} & \makecell[c]{\CheckmarkBold}     & \makecell[c]{\CheckmarkBold}     & \makecell[c]{\CheckmarkBold}     & \makecell[c]{\XSolidBrush}     & \makecell[c]{\XSolidBrush} \\
    \multicolumn{1}{c}{} & \makecell[c]{Enhanced Prompt\\Strategy} & \makecell[c]{\CheckmarkBold}     & \makecell[c]{\CheckmarkBold}     & \makecell[c]{\CheckmarkBold\kern-1.2ex\raisebox{1ex}{\rotatebox[origin=c]{125}{\textbf{--}}}} & \makecell[c]{\XSolidBrush}     & \makecell[c]{\XSolidBrush} \\
    \multicolumn{1}{c}{} & \makecell[c]{Detector-Based} & \makecell[c]{\CheckmarkBold}     & \makecell[c]{\CheckmarkBold}     & \makecell[c]{\CheckmarkBold}     & \makecell[c]{\XSolidBrush}     & \makecell[c]{\CheckmarkBold} \\
    \midrule
    \multirow{4}[1]{*}{\makecell[c]{In-Decoding\\TF Alignment}} & \makecell[c]{Hidden States\\Adjustment} & \makecell[c]{\XSolidBrush}     & \makecell[c]{\XSolidBrush}     & \makecell[c]{\CheckmarkBold}     & \makecell[c]{\XSolidBrush}     & \makecell[c]{\CheckmarkBold} \\
    \multicolumn{1}{c}{} & \makecell[c]{Logits Difference\\Calculation} & \makecell[c]{\XSolidBrush}     & \makecell[c]{\XSolidBrush}     & \makecell[c]{\CheckmarkBold}     & \makecell[c]{\XSolidBrush}     & \makecell[c]{\CheckmarkBold}\\
    \multicolumn{1}{c}{} & \makecell[c]{Guidance-Based} & \makecell[c]{\XSolidBrush}     & \makecell[c]{\XSolidBrush}     & \makecell[c]{\CheckmarkBold}     & \makecell[c]{\XSolidBrush}     & \makecell[c]{\CheckmarkBold} \\
    \multicolumn{1}{c}{} & \makecell[c]{Dynamic Search\\Strategy} & \makecell[c]{\XSolidBrush}     & \makecell[c]{\XSolidBrush}     & \makecell[c]{\XSolidBrush}     & \makecell[c]{\XSolidBrush}     & \makecell[c]{\CheckmarkBold} \\
    \midrule
    \makecell[c]{Post-Decoding\\TF Alignment} & \makecell[c]{Filtering/Correcting\\Outputs} & \makecell[c]{\CheckmarkBold}     & \makecell[c]{\CheckmarkBold}     & \makecell[c]{\CheckmarkBold}     & \makecell[c]{\XSolidBrush}     & \makecell[c]{\CheckmarkBold} \\
    \bottomrule
    \end{tabular}%
  }
  \caption{\label{tab:comparison2}
  A Comparative Analysis of Different TF Alignment Methods. (Note: \CheckmarkBold and \XSolidBrush denote the method has/hasn’t the corresponding property. \CheckmarkBold\kern-1.2ex\raisebox{1ex}{\rotatebox[origin=c]{125}{\textbf{--}}} denotes differential storage requirements, where enhanced prompt strategies such as RAG-based architectures or Chain-of-Thought augmented frameworks necessitate storage allocation, whereas other methods remain storage-free.)
  }
\end{table*}%

Taking safety alignment as an example, the analysis in Table~\ref{tab:numerical2} compares the safety alignment performance of different methods applied to llama2-7b-chat. 
Detailed information regarding datasets, evaluation criteria, results analysis, and comparisons of various TF alignment methods is presented as follows.

\subsection{Dataset and Evaluation Criteria}
Following SCANS~\cite{cao2025scans}, we selected AdvBench and TruthfulQA~\cite{Lin2021TruthfulQAMH} datasets to evaluate the safety and helpfulness of aligned models. Building on this, we incorporated the more challenging SafeEdit~\cite{wang-etal-2024-detoxifying} dataset to assess how the aligned models perform against adversarial attacks.
AdvBench consists of 456 harmful queries. SafeEdit, comprising 500 adversarial attacks across 9 unsafe scenarios, is used to evaluate the robustness of these alignment methods under extreme adversarial conditions.
Besides, TruthfulQA is used as the test set for evaluating helpfulness, consisting of 753 benign questions. 

We comprehensively assess the effectiveness of different alignment methods from two crucial dimensions: the Defense Success Rate, which reflects safety, and the Benign Refusing Rate, which reflects helpfulness.
Regarding the Defense Success Rate, we employ the safety judgement classifier proposed by ~\citet{wang-etal-2024-detoxifying} to evaluate whether a response is safe.
When evaluating the Benign Refusing Rate, we utilize the classic keyword detection method. If responses to safety questions following the application of an alignment method contain more refusal-related keywords, the method demonstrates lower utility.

\subsection{More Numerical Analysis}

Data from Table~\ref{tab:numerical2} reveals that TF alignment methods can rival or even surpass the safety and helpfulness performance of FT alignment methods. For example, while SCANS demonstrates safety performance comparable to that of FT-based SafeDecoding, it exhibits superior usefulness.
Furthermore, all TF alignment methods presented in the table demonstrate higher levels of helpfulness.
These findings underscore that TF alignment methods merit further investigation and refinement, particularly in contexts where FT alignment methods face constraints, such as severe knowledge distortion and model accessibility.

Specifically, URIAL demonstrates poor performance on the SafeEdit dataset because the method uses only three examples as prompts, which are fed into the LLM alongside the question. This approach fails to generalize well to more challenging adversarial attack problems, as discussed in Section~\ref{sec_Discussion_1}.
Additionally, comparing the performance of different methods on TruthfulQA can partially reflect the degree of impairment to the model’s inherent capabilities. It is evident that FT alignment methods cause the most severe knowledge impairment, while TF alignment methods exhibit a milder impact by comparison. For further comparison, experiments can be conducted on datasets such as MMLU and XSum (as discussed in Section~\ref{section_concepts_evaluate}).

\subsection{When and How to Choose TF Alignment?} 
\label{section_concepts_criteria}

We conclude that TF alignment methods are particularly well-suited for the following scenarios: (1) \textit{Resource-Constrained Settings}: When computational resources or training data are limited, TF alignment methods offer a lightweight and cost-effective solution. (2) \textit{Black-Box Models}: When the model parameters are inaccessible (e.g., closed-source or proprietary models), TF alignment methods provide a viable alternative to FT. (3) \textit{Rapid Deployment}: When quick adaptation is required, TF alignment methods like in-context learning or decoding-time adjustments enable immediate alignment without lengthy FT processes. (4) \textit{Preserving Knowledge}: When FT alignment risks the degrading the model’s pre-existing knowledge, TF alignment methods mitigate this issue. 
(5) \textit{Stylistic or Normative Alignment}: 
Empirical evidence from URIAL demonstrates that when the goal is to influence stylistic elements, e.g., tone, politeness, or to ensure ethical compliance without altering factual content, TF alignment can achieve comparable or superior performance of FT alignment purely through ICL with base LLMs.

However, this does not mean that one method can be applied to all the above scenarios, for every method has its own strengths and limitations. Therefore, Table~\ref{tab:comparison2} systematically compares the underlying characteristics, compatibility (model accessibility and plug-and-play capability), storage and real-time efficiency, and generalization of different TF alignment methods.
We can observe that all in-decoding TF alignment methods are inapplicable to closed-source LLMs, yet they exhibit strong generalization capability. Additionally, TF alignment methods introduce latency due to increased prompt length, inference-time adjustments, or input-output detection, resulting in suboptimal real-time efficiency.

We hope these comparisons can enable practitioners to gain a nuanced understanding of each approach and make well-informed decisions when selecting alignment strategies.

\end{document}